\pdfoutput=1
\documentclass[11pt]{article}

\usepackage[]{acl}

\usepackage{times}
\usepackage{latexsym}

\usepackage[T1]{fontenc}
\usepackage[utf8]{inputenc}

\usepackage{microtype}

\title{Text-to-Table: A New Way of Information Extraction}

\usepackage[switch]{lineno}
\usepackage{nicematrix}
\usepackage{amssymb}
\usepackage{pifont}
\usepackage{subcaption}
\usepackage[linesnumbered,ruled,vlined,algo2e]{algorithm2e}
\usepackage{enumitem}
\usepackage{xparse}
\usepackage{multirow}
\usepackage{multicol}

\newcolumntype{I}{!{\OnlyMainNiceMatrix{\vrule}}}

\NewDocumentCommand{\xueqing}
{ mO{} }{\textcolor{cyan}{\textsuperscript{\textit{Xueqing}}\textsf{\textbf{\small[#1]}}}}

\newcommand{\cmark}{\ding{51}}%
\newcommand{\xmark}{\ding{55}}%

\newcommand{\sep}{\text{$\langle \textmd{s} \rangle$}}
\newcommand{\nline}{\text{$\langle \textmd{n} \rangle$}}

\newcommand{\eos}{\text{$\langle \textmd{eos} \rangle$}}

\author{Xueqing Wu\textsuperscript{1}, Jiacheng Zhang\textsuperscript{2}, Hang Li\textsuperscript{2} \\
  University of Illinois Urbana-Champaign\textsuperscript{1} \\
  ByteDance AI Lab\textsuperscript{2} \\
  \texttt{xueqing8@illinois.edu} \\
   \texttt{{zhangjiacheng.grit,lihang.lh}@bytedance.com} }

\begin{document}

\newcommand\mycommfont[1]{\footnotesize\textcolor{blue}{#1}}
\SetCommentSty{mycommfont}
\SetFuncSty{text} % default: texttt
\SetKwInput{KwInput}{Input}                % Set the Input
\SetKwInput{KwOutput}{Output}              % set the Output
\SetKwRepeat{Do}{do}{while}

% \makeatletter
% \newcommand{\nosemic}{\renewcommand{\@endalgocfline}{\relax}}% Drop semi-colon ;
% \newcommand{\dosemic}{\renewcommand{\@endalgocfline}{\algocf@endline}}% Reinstate semi-colon ;
% \newcommand{\pushline}{\Indp}% Indent
% \newcommand{\popline}{\Indm\dosemic}% Undent
% \let\oldnl\nl% Store \nl in \oldnl
% \newcommand{\nonl}{\renewcommand{\nl}{\let\nl\oldnl}}% Remove line number for one line
% \makeatother

% \linenumbers

\maketitle

\begin{abstract}
We study a new problem setting of information extraction (IE), referred to as text-to-table. In text-to-table, given a text, one creates a table or several tables expressing the main content of the text, while the model is learned from text-table pair data. The problem setting differs from those of the existing methods for IE. First, the extraction can be carried out from long texts to large tables with complex structures. Second, the extraction is entirely data-driven, and there is no need to explicitly define the schemas.  As far as we know, there has been no previous work that studies the problem. In this work, we formalize text-to-table as a sequence-to-sequence (seq2seq) problem.  We first employ a seq2seq model fine-tuned from a pre-trained language model to perform the task.  We also develop a new method within the seq2seq approach, exploiting two additional techniques in table generation: table constraint and table relation embeddings. We consider text-to-table as an inverse problem of the well-studied table-to-text, and make use of four existing table-to-text datasets in our experiments on text-to-table. Experimental results show that the vanilla seq2seq model can outperform the baseline methods of using relation extraction and named entity extraction. The results also show that our method can further boost the performances of the vanilla seq2seq model. We further discuss the main challenges of the proposed task. The code and data are available at \url{https://github.com/shirley-wu/text_to_table}. \footnote{The work was done when Xueqing Wu was an intern at ByteDance AI Lab.}
\end{abstract}

\section{Introduction}

\begin{figure}[tbh]
\centering
\includegraphics[width=0.95\linewidth]{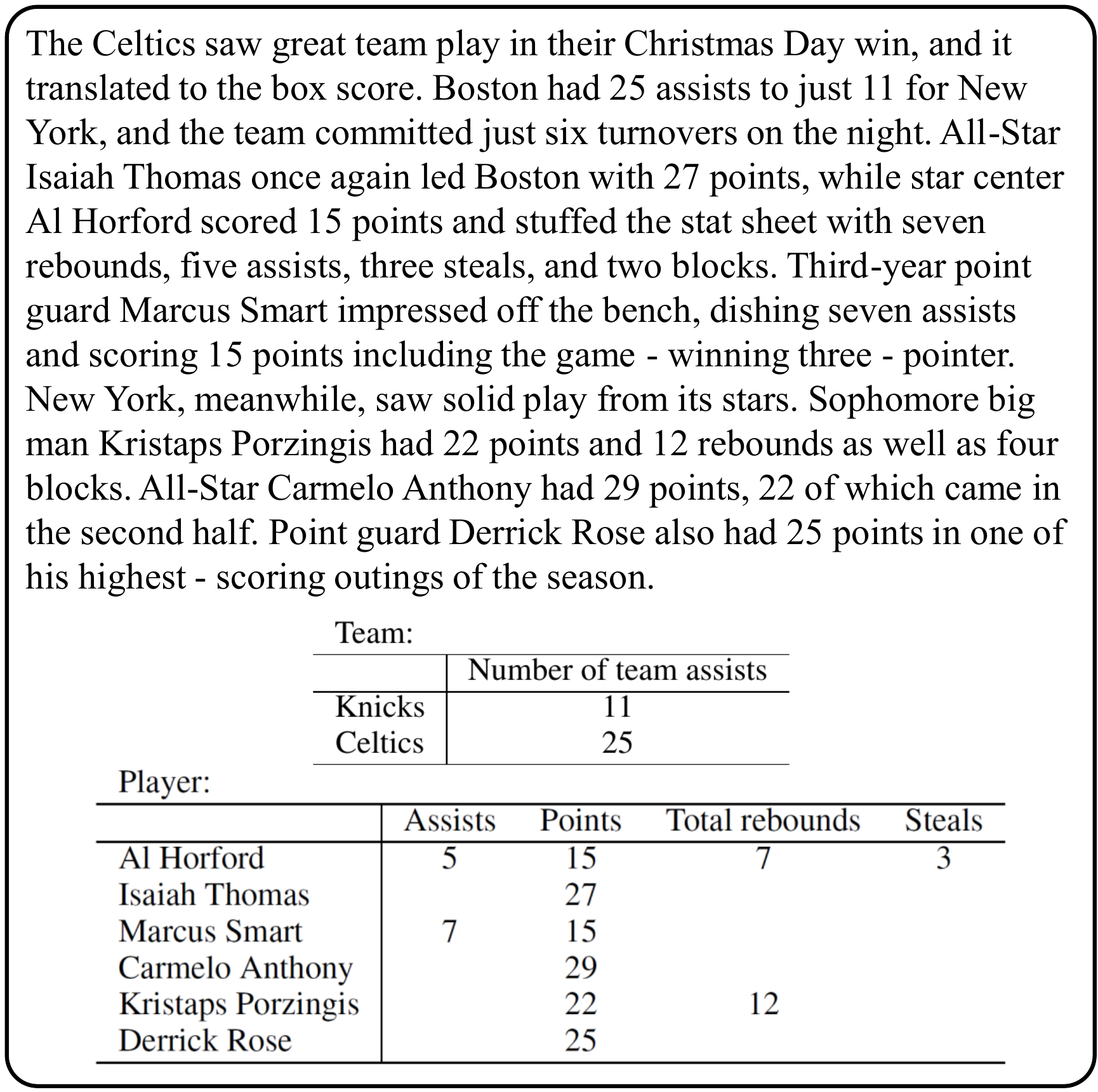}
\caption{An example of text-to-table from the Rotowire dataset. The text is a report of a basketball game, and the tables are the scores of the teams and players.}
\label{fig:rotowire_example}
\end{figure}
 
Information extraction (IE) is a task that aims to extract information of interest from text data and represent the extracted information in a structured form.
Traditional IE tasks include named entity recognition which recognizes entities and their types \citep{huang2015bidirectional,ma2016end,lample2016neural,devlin2019bert}, relation extraction which identifies the relationships between entities \citep{zheng2017joint,zeng2018extracting,luan2019general,zhong-chen-2021-frustratingly}, etc. Since the results of IE are structured, they can be easily used by computer systems in different applications such as text mining.

In this work, we study IE in a new setting, referred to as text-to-table. First, the
system receives a training dataset containing text-table pairs.  Each text-table pair contains a text and a table (or tables) representing information needed for the target application extracted from the text.  The system learns a model for information extraction.  Next, the system employs the learned model to conduct information extraction from a new text and outputs the result in a table (or tables). Figure \ref{fig:rotowire_example} gives an example of text-to-table, where the input (above) is a report of a basketball game, and the output (below) is two tables summarizing the scores of the teams and players from the input.

Text-to-table is unique compared to the traditional IE approaches.
First, text-to-data can be performed at both sentence-level and document-level.
While the distinction between sentence and document level is vague, document-level extraction can produce a more complex output.
As in the example in Figure \ref{fig:rotowire_example}, extraction of information is performed from the entire document. The extracted information contains multiple types of scores of teams and players in a basketball game structured in table format. % This is different from traditional IE methods such as named entity recognition or relation extraction that extract multiple independent entities or relations.
% First, it is mainly designed to extract information on complex relations between items from a long text (e.g., the whole document).
Second, the schemas for extraction are implicitly included in the training data such as header names. There is no need to explicitly define the schemas, which reduces the need for manual efforts for schema design and annotations.

% Given the differences, we cannot directly use existing IE datasets for this setting.
Our work is inspired by research on the so-called table-to-text (or data-to-text) problem, which is the task of generating a description for a given table. Table-to-text is useful in applications where the content of a table needs to be described in natural language. Thus, text-to-table can be regarded as an inverse problem of table-to-text.  However, there are also differences. Most notably, their applications are different. % Text-to-table can be applied to document summarization, text mining, etc.
Text-to-table systems can automatically produce tables for text summarization and text mining. For example, the score tables of sports games and infoboxes of Wikipedia articles can serve as summaries of original documents.
%Compared to text summaries, table-format summaries present the salient information in a structured way and can help readers' understanding. \xueqing{Is this overclaim?}
The score tables can be utilized to evaluate the athletes' performances, and the infoboxes can be used to construct a knowledge graph.

In this work, we formalize text-to-table as a sequence-to-sequence (seq2seq) task. More specifically, we translate the text into a sequence representation of a table (or tables), where the schema of the table is implicitly contained in the representation. 
We build the seq2seq model on top of a pre-trained language model, which is the state-of-the-art approach for seq2seq tasks \citep{DBLP:conf/acl/LewisLGGMLSZ20,raffel2020exploring}.
%We also assume that the seq2seq model is built on top of a pre-trained language model, such as BART~\citep{DBLP:conf/acl/LewisLGGMLSZ20} and T5~\cite{raffel2020exploring}. 
%(Note that BART is more suitable for models based on Transformer encoder-decoder, while BERT is more suitable for models based on Transformer encoder.) 
Although the approach is a natural application of existing technologies, as far as we know, there has been no previous study to investigate to what extent the approach works. We also develop a new method for text-to-table within the seq2seq approach with two additional techniques, table constraint, and table relation embeddings. Table constraint controls the creation of rows in a table and table relation embeddings affect the alignments between cells and their row headers and column headers. Both are to make the generated table well-formulated.

The approach to IE based on seq2seq has already been proposed. Methods for conducting individual tasks of relation extraction~\citep{zeng2018extracting,nayak2020effective,DBLP:conf/emnlp/HuangTP21}, 
%entity linking~\citep{DBLP:conf/iclr/CaoI0P21}, 
named entity recognition~\citep{chen2018learning,DBLP:conf/acl/YanGDGZQ20}, event extraction~\citep{DBLP:conf/naacl/LiJH21,DBLP:conf/acl/0001LXHTL0LC20} and role-filler entity extraction \citep{DBLP:conf/eacl/DuRC21,DBLP:conf/emnlp/HuangTP21} have been developed. Methods for jointly performing multiple tasks of named entity recognition, relation extraction, and event extraction have also been devised~\citep{DBLP:conf/iclr/PaoliniAKMAASXS21}.  Most of the methods exploit suitable pre-trained models such as BERT. However, all the existing methods rely on pre-defined schemas for extraction. Moreover, their models are designed to extract information from short texts, rather than long texts, and extract information with simple structures (such as an entity and its type), rather than information with complicated structures (such as a table). %simply structured information, not complicatedly structured information.

We conduct extensive experiments on the four datasets. Results show that the vanilla seq2seq model fine-tuned from BART~\citep{DBLP:conf/acl/LewisLGGMLSZ20} can outperform the state-of-the-art IE models fine-tuned from BERT~\citep{devlin2019bert,zhong-chen-2021-frustratingly}. Furthermore, results show that our proposed approach to text-to-table with the two techniques can further improve the extraction accuracies. We also summarize the challenging issues with the seq2seq approach to text-to-table for future research.

Our contributions are summarized as follows:
\begin{enumerate}[noitemsep,partopsep=0pt,topsep=0pt,parsep=0pt]
    \item We propose the new task of text-to-table for IE. We derive four new datasets for the task from existing datasets.
    \item We formalize the task as a seq2seq problem and propose a new method within the seq2seq approach using the techniques of table constraint and table relation embeddings.
    \item We conduct extensive experiments to verify the effectiveness of the proposed approach.
\end{enumerate}

\section{Related Work}

\textbf{Information Extraction} (IE) is the task of extracting information  (structured data) from a text (unstructured data).
%, and there are some specific cases. 
For example, named entity recognition (NER) recognizes entities appearing in a text. Relation extraction (RE) identifies the relationships between entities. Event extraction (EE) discovers events occurring in a text. Role-filler entity extraction (REE) fills entities into event templates and is similar to EE.

Traditionally, researchers formalize the task as a language understanding problem. The state-of-the-art methods for NER perform the task on the basis of the pre-trained language model BERT~\citep{devlin2019bert}. The pipeline approach to RE divides the problem into NER and relation classification, and conducts the two sub-tasks in a sequential manner~\citep{zhong-chen-2021-frustratingly}, while the end-to-end approach jointly carries out the two sub-tasks~\citep{zheng2017joint,zeng2018extracting,luan2019general}. 
The state-of-the-art methods for EE also employ BERT and usually jointly train the models with other tasks such as NER and RE~\citep{wadden2019entity,zhang2019joint,lin2020joint}.
All the methods assume the use of pre-defined schemas (e.g., entity types for NER, entity and relation types for RE, and event templates for EE). Besides, most methods are designed for extraction from short texts. Therefore, existing methods for IE cannot be directly applied to text-to-table.

Another series of related work is open information extraction (OpenIE), which aims to extract information from texts without relying on explicitly defined schemas \citep{DBLP:conf/ijcai/BankoCSBE07,DBLP:conf/acl/WuW10,DBLP:conf/emnlp/MausamSSBE12,DBLP:conf/naacl/StanovskyMZD18,DBLP:conf/aaai/ZhanZ20}.
However, OpenIE aims to extract information with simple structures (i.e., relation tuples) from short texts, and the methods in OpenIE cannot be directly applied to text-to-table.

IE is also conducted at document level, referred to as doc-level IE.
For example, some NER methods directly perform NER on a long document~\citep{strubell2017fast,luo2018attention}, and others encode each sentence in a document, use attention to fuse document-level information, and perform NER on each sentence~\citep{hu2020leveraging,xu2018improving}. There are also RE methods that predict the relationships between entities in a document~\citep{yao2019docred,nan2020reasoning}. However, existing doc-level IE approaches usually do not consider the extraction of complex relations between many items.

\textbf{Sequence-to-sequence} (seq2seq) is the general problem of transforming one text into another text~\cite{sutskever2014sequence,DBLP:journals/corr/BahdanauCB14}, which includes machine translation, text summarization, etc. The use of the pre-trained language models of BART~\citep{DBLP:conf/acl/LewisLGGMLSZ20} and T5~\cite{raffel2020exploring} can significantly boost the performances of seq2seq, such as machine translation \citep{DBLP:conf/acl/LewisLGGMLSZ20,raffel2020exploring,liu2020multilingual} and text summarization \citep{DBLP:conf/acl/LewisLGGMLSZ20,raffel2020exploring,huang2020have}.

Recently, some researchers also formalize the IE problems as seq2seq, that is, transforming the input text into an internal representation. One advantage is that one can employ a single model to extract multiple types of information.
Results show that this approach works better than or equally well as the traditional approach of language understanding, in RE~\citep{zeng2018extracting,nayak2020effective}, NER~\citep{chen2018learning,DBLP:conf/acl/YanGDGZQ20}, EE~\citep{DBLP:conf/naacl/LiJH21,DBLP:conf/acl/0001LXHTL0LC20} and REE \citep{DBLP:conf/eacl/DuRC21,DBLP:conf/emnlp/HuangTP21}.
Methods that jointly perform multiple tasks including NER, RE, and EE have also been devised \citep{DBLP:conf/iclr/PaoliniAKMAASXS21}.

\textbf{Data-to-text} aims to generate natural language descriptions from the input structured data such as sports commentaries~\cite{wiseman2017challenges}. %,thomson2020sportsett}, restaurant reviews \cite{novikova2017e2e}, biographical texts~\cite{lebret2016neural}, and open-domain descriptions \cite{parikh2020totto,DBLP:conf/naacl/NanRZRSHTVVKLIP21}.
The structured data is usually represented as tables~\cite{wiseman2017challenges,thomson2020sportsett,chen2020logical}, sets of table cells~\cite{parikh2020totto,bao2018table}, semantic representations~\cite{novikova2017e2e}, or sets of relation triples~\cite{gardent2017creating,DBLP:conf/naacl/NanRZRSHTVVKLIP21}.
The task requires the model to select the salient information from the data, organize it in a logical order, and generate an accurate and fluent natural language description~\cite{wiseman2017challenges}.  Data-to-text models usually adopt the encoder-decoder architecture. 
The encoders are specifically designed to model the input data, such as multi-layer perceptron \citep{puduppully2019data,puduppully2019data2}, recurrent neural network \citep{juraska2018deep,liu2018table,shen2020neural}, graph neural network \citep{marcheggiani2018deep,koncel2019text}, or Transformer \citep{gong2019enhanced}.

\section{Problem Formulation}

% \subsection{Task Formulation}

As shown in Figure \ref{fig:rotowire_example}, text-to-table takes a text as input and produces a table or several tables to summarize the content of the text. %, as shown in Figure 1. This can be considered as an inverse problem of data-to-text, and each has its applications.

Formally, the input is a text denoted as $\mathbf{x}=x_1,\cdots,x_{|\mathbf{x}|}$. The output is one table or multiple tables. For simplicity suppose that there is only one table denoted as $T$. Further, suppose that $T$ has $n_r$ rows and $n_c$ columns. Thus, $T$ contains $n_r \times n_c$ cells, where the cell of row $i$ and column $j$ is a sequence of words $\mathbf{t}_{i,j} = t_{i,j,1}, ..., t_{i,j,|\mathbf{t}_{i,j}|}$.

There are three types of table: one that has both column headers and row headers, one that only has column headers, and one that only has row headers.  For example, the player table in Figure \ref{fig:rotowire_example} has both column headers (``Assists'', ``Points'', etc) and row headers (``Al Horford'', ``Isaiah Thomas'', etc).  We let $\mathbf{t}_{1,j},\ j = 2, \cdots, n_c$ denote the column headers, let $\mathbf{t}_{i,1},\ i = 2, \cdots, n_r$ denote the row headers, and let $\mathbf{t}_{i,j},\ i = 2, \cdots, n_r, j = 2, \cdots, n_c$ denote the non-header cells of the table. For example, in the player table in Figure \ref{fig:rotowire_example}, $\mathbf{t}_{1,2}=\text{Assists}$,  $\mathbf{t}_{2,1}=\text{Al Horford}$, and $\mathbf{t}_{2,2}=\text{5}$.
%\red{[del a para]}

\iftrue
% We consider using machine learning to perform text-to-table. In learning, a number of text-table pairs are given as training data, and a model is trained from the data.  In inference, the learned model is utilized to generate a table or tables given a new text.

% Once the information of a text is extracted into tables via text-to-table, it can be leveraged in many different applications such as document summarization and text mining.
The information extracted via text-to-table can be leveraged in many different applications such as document summarization and text mining. For example, in Figure \ref{fig:rotowire_example}, one can quickly obtain the key information of the text by simply looking at the tables summarized from the text.

There are differences between text-to-table and traditional IE settings. As can be seen from the example in Figure \ref{fig:rotowire_example}, extraction of information is performed from the entire document. The extracted information (structured data) is in a complex form, specifically multiple types of scores of teams and players in a basketball game. Furthermore, the data-driven approach is taken, and the schemas of the tables do not need to be explicitly defined.
\fi

The task of text-to-table also has challenges.
First, parallel data containing texts and tables is difficult to obtain. Manual construction of such data is usually expensive.
Second, structured information may not be easily represented as tables. For example, a knowledge graph may not be easily converted into tables. 
Third, evaluation of table extraction may not be easy, which includes multiple factors, such as header, content, and structure. 

%Furthermore, in open-vocabulary scenarios such as Wikipedia, text-to-table often involves inflexion (e.g. winner v.s. winners) and paraphrasing (e.g., england v.s. english), which are difficult to evaluate automatically.

\section{Our Method}

We develop a method for text-to-table using the seq2seq approach and the two techniques of table constraint and table relation embeddings.

\subsection{Vanilla Seq2Seq} % Framework} 

\begin{figure}[htb]
\centering
\includegraphics[width=\linewidth]{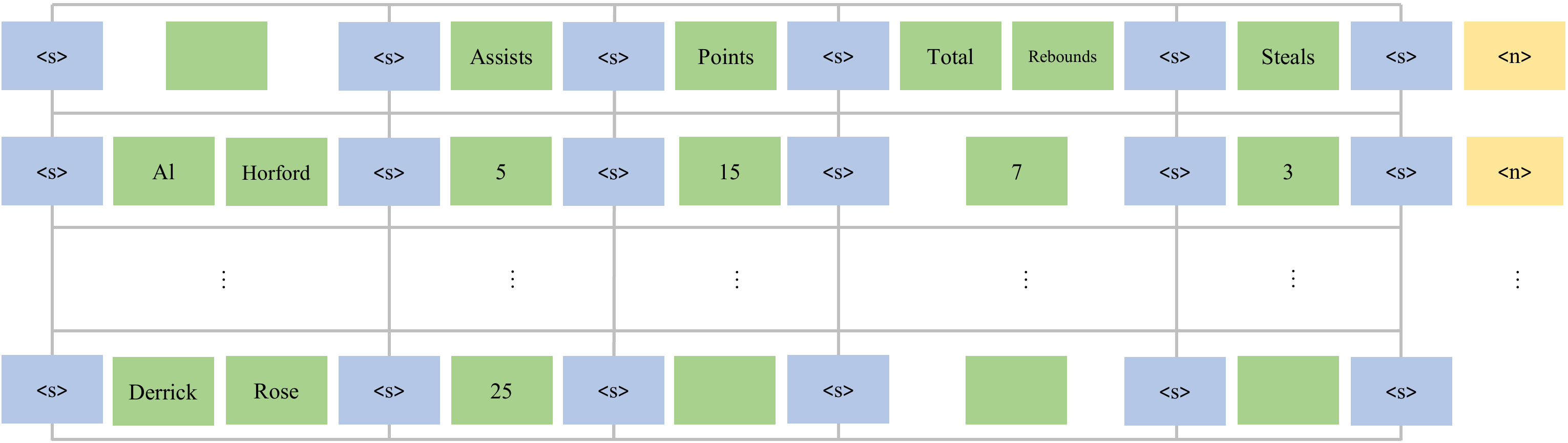}
\caption{The sequence representation of the player table in Figure \ref{fig:rotowire_example}. %The green blocks represent normal text tokens, 
The blue items are separation tokens \sep{} and the yellow items are new-line tokens \nline{}.}
\label{fig:text_repr}
\end{figure}

We formalize text-to-table as a sequence-to-sequence (seq2seq) problem~\citep{sutskever2014sequence,DBLP:journals/corr/BahdanauCB14}. Specifically, given an input text, we generate a sequence representing the output table (or tables). 
We introduce two special tokens, a separation token denoted as “\sep{}” and a new-line token denoted as “\nline{}”.
For a table $\mathbf{t}$, we represent each row $\mathbf{t}_{i}$ with a sequence of cells delimited by separation tokens:
\begin{align}
\mathbf{t}_{i} = \sep{},\mathbf{t}_{i,1},\sep{},\cdots,\sep{},\mathbf{t}_{i,n_c},\sep{}.
\end{align}
We represent the entire table with a sequence of rows delimited by new-line tokens:
\begin{align}
\mathbf{t} & = &  \sep{},\mathbf{t}_{1,1},\sep{},\cdots,\sep{},\mathbf{t}_{1,n_c},\sep{} ,\nline{}, \\ \nonumber
 &  & \sep{},\mathbf{t}_{2,1},\sep{},\cdots,\sep{},\mathbf{t}_{2,n_c},\sep{} ,\nline{}, \\ \nonumber
 & & \cdots \cdots \\ \nonumber
& & \sep{},\mathbf{t}_{n_r,1},\sep{},\cdots,\sep{},\mathbf{t}_{n_r,n_c},\sep{} \quad \nonumber
\end{align}

Figure~\ref{fig:text_repr} shows the sequence of the player table in Figure~\ref{fig:rotowire_example}. When there are multiple tables, we create a sequence of tables using the captions of the tables as delimiters.

Let $\mathbf{x}=x_1,\cdots,x_{|\mathbf{x}|}$ and $\mathbf{y}=y_1,\cdots,y_{|\mathbf{y}|}$ denote the input and output sequences respectively. In inference, the model generates the output sequence based on the input sequence. The model conducts generation in an auto-regressive way, which generates one token at each step based on the tokens it has generated so far. In training, we learn the model based on the text-table pairs $\{(\mathbf{x}_1,\mathbf{y}_1),(\mathbf{x}_2,\mathbf{y}_2),\cdots, (\mathbf{x}_n,\mathbf{y}_n)\}$. The objective of learning is to minimize the cross-entropy loss.

\iffalse
Let $\mathbf{x}=x_1,\cdots,x_{|\mathbf{x}|}$ and $\mathbf{y}=y_1,\cdots,y_{|\mathbf{y}|}$ denote the input and output sequences respectively. In inference, the model generates the output sequence based on the input sequence. We conduct decoding in an auto-regressive way, which generates one token at each step based on the tokens it has generated so far. Formally, the model  calculates the conditional probability:
\begin{align}
  p(\mathbf{y} | \mathbf{x}) =  \prod_{t=1}^{|\mathbf{y}|} p(y_i|\mathbf{x},\mathbf{y}_{<t})
\end{align}
We use decoding algorithms such as beam search or greedy search to find an output sequence that approximately maximizes the conditional probability.

% \begin{align}
% \arg \max_{\mathbf{y}} p(\mathbf{t} | \mathbf{x}) = \arg \max_{\mathbf{y}} \prod_{t=1}^{|\mathbf{y}|} p(y_t|\mathbf{x},\mathbf{y}_{<t})
% \end{align}

In training, we learn the model based on the text-table pairs $\{(\mathbf{x}_1,\mathbf{y}_1),(\mathbf{x}_2,\mathbf{y}_2),\cdots, (\mathbf{x}_n,\mathbf{y}_n)\}$. The objective of learning is to minimize the cross-entropy loss
\begin{align}
\arg \min_{\mathbf{\theta}} L = \arg \min_{\mathbf{\theta}} - \sum_{i=1}^{n} \log p_{\mathbf{\theta}}(\mathbf{y}_i|\mathbf{x}_i)
\end{align}
where $\mathbf{\theta}$ denotes the parameter of the model.

In our work, we adopt Transformer as the model~\citep{vaswani2017attention}, which is the state-of-the-art method for seq2seq. We build the model on top of the pre-trained language model BART~\citep{DBLP:conf/acl/LewisLGGMLSZ20} by fine-tuning.
\fi

We refer to the method described above as ``vanilla seq2seq''. There is no guarantee, however, that the output sequence of vanilla seq2seq represents a well-formulated table. We add a post-processing step to ensure that the output sequence is a table. The post-processing method takes the first row generated as well-defined, deletes extra cells at the end of the other rows, and inserts empty cells at the end of the other rows.

\subsection{Techniques}

We develop two techniques to improve table generation, called table constraint and table relation embeddings. We use ``our method'' to denote the seq2seq approach with these two techniques.\footnote{Our methods is able to generate the output containing multiple tables. This is discussed in Appendix \ref{sec:appendix:multitable}.} %Both techniques are designed to improve table generation.
%\red{For simplicity, we only discuss the case where the output contains only one table. Our methods when the output contains multiple tables are discussed in Appendix \ref{sec:appendix:multitable}.}

\subsubsection*{Table Constraint} 

Our method exploits a constraint in the decoding process to ensure that the output sequence represents a well-formulated table. Specifically, our method calculates the number of cells in the first row it generates, and then forces the following rows to contain the same number of cells.
%\red{The pseudo-codes are in Algorithm \ref{alg:constrained_decoding} in Appendix \ref{sec:appendix:constrained_decoding}. [del algo]}
% 
% First, the decoder generates the first row (line \ref{alg:constrained_decoding:repeat_first_row_start} - \ref{alg:constrained_decoding:repeat_first_row_end}), and it completes the process by generating either an end-of-sentence token \eos{}\footnote{In case the table contains only one row.} or a new-line token \nline{} after a separation token \sep{}. Next, the decoder calculates the number of cells in the first row $n_c$ (line \ref{alg:constrained_decoding:calc_nc}).
% After that, the decoder generates the next row (line 14 - 21) by generating $n_c$ cells delimited by \sep{}, followed by either \eos{} or \nline{}.
% The decoder repeats the process until it generates \eos{}.

\subsubsection*{Table Relation Embeddings}

Our method also incorporates table relation embeddings including row relation embeddings and column relation embeddings into the self-attention of the Transformer decoder.
Given a token in a non-header cell, the row relation embeddings $\tau_r^K$ and $\tau_r^V$ indicate which row header the token is aligned to, and the column relation embeddings $\tau_c^K$ and $\tau_c^V$ indicate which column header the token is aligned to.
% Given a non-header cell, the embeddings of row header \todo{} indicate whether the cell and the row header are on the same row, and given a non-header cell, the embeddings of column header indicate whether the cell and the column header are on the same column.

Let us consider the self-attention function in one block of Transformer decoder: at each position, self-attention only attends to the previous positions. For simplicity, let us only consider one head in the self-attention. 
At the $t$-th position, the input of self-attention is the sequence of representations $z=(z_1,\cdots,z_t)$ and the output is the sequence of representations $h=(h_1,\cdots,h_t)$, where $z_i\in \mathbb{R}^{d}$ and $h_i\in \mathbb{R}^{d}$ are the representations at the $i$-th position $(i=1,\cdots, t)$.

In a conventional Transformer decoder, self-attention is defined as follows,
\begin{align}
& h_i = \left(\sum_{j=1}^i \alpha_{ij} (z_j W^V)\right) W^O, \\
& \alpha_{ij} = \frac{e^{e_{ij}}}{\sum_{j=1}^i e^{e_{ij}}}, e_{ij}=\frac{(z_i W^Q)(z_j W^K)^{\text{T}}}{\sqrt{d_k}}, \\
& i=1,\cdots,t,\ j=1,\cdots, i \nonumber
\end{align}
where $W^Q,W^K,W^V\in \mathbb{R}^{d \times d_k}$ are the query, key, and value weight matrices respectively, and $W^O\in \mathbb{R}^{d_k \times d}$ is the output weight matrix. 

In our method, self-attention is defined as:
\begin{align}
& h_i = \left(\sum_{j=1}^{i} \alpha_{ij} (z_j W^V + r_{ij}^V)\right) W^O, \\
& \alpha_{ij} = \frac{e^{e_{ij}}}{\sum_{j=1}^i e^{e_{ij}}}, e_{ij} = \frac{(z_i W^Q) (z_j W^K + r_{ij}^K)^{\text{T}}}{\sqrt{d_k}}, \\
& i=1,\cdots,t,\ j=1,\cdots, i \nonumber
\end{align}
\begin{figure}[htb]
\centering
\includegraphics[width=\linewidth]{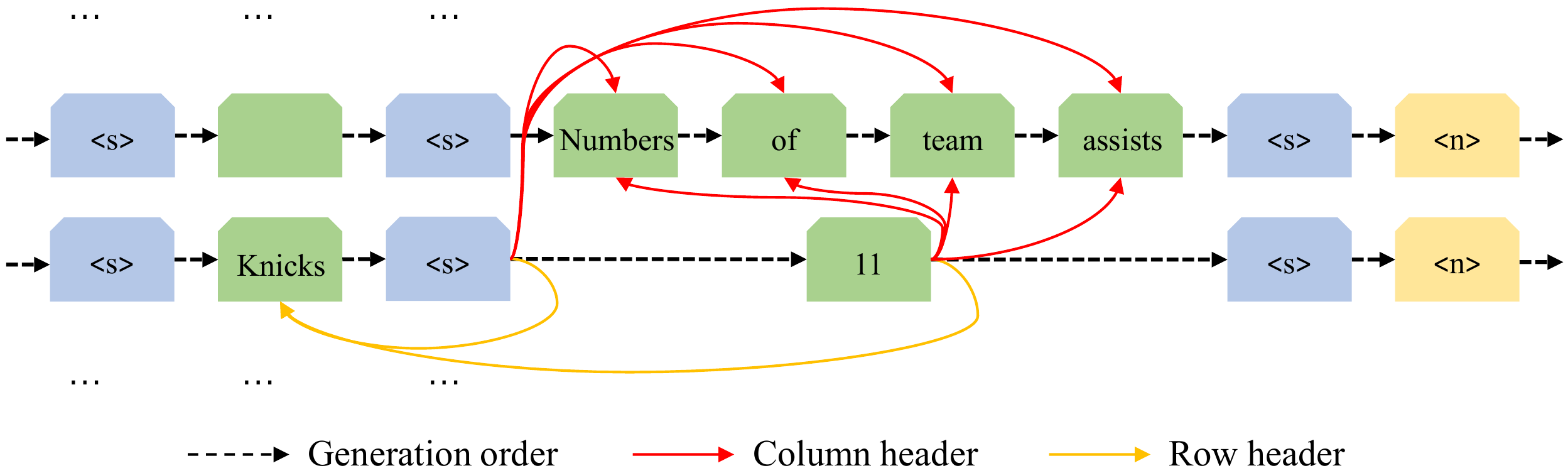}
\caption{Construction of relation vectors. Red and yellow arrows represent alignments with column headers and row headers respectively. The relation vectors regarding tokens ``11'' and one \sep{} are illustrated.}
% we need to have a consistent usage for quotations
\label{fig:relation}
\end{figure}
where $r_{ij}^K$ and $r_{ij}^V$ are relation vectors representing the relationship between the $i$-th position and the $j$-th position.

The relation vectors $r_{ij}^K$ and $r_{ij}^V$ are defined as follows. For the token at the $i$-th position, if the token at the $j$-th position is a part of its row header, then $r_{ij}^K$ and $r_{ij}^V$ are set to the row relation embeddings $\tau_r^K$ and $\tau_r^V$. Similarly, for the token at the $i$-th position, if the token at the $j$-th position is a part of its column header, then $r_{ij}^K$ and $r_{ij}^V$ are set to the column relation embeddings $\tau_c^K$ and $\tau_c^V$. Otherwise, $r_{ij}^K$ and $r_{ij}^V$ are set to $0$. In inference, to identify the row header or the column header of a token, we parse the sequence generated so far to create a partial table using the new-line tokens and separation tokens in the sequence.
Figure \ref{fig:relation} illustrates how relation vectors are constructed.

\begin{table*}[!htb]
\centering
\begin{tabular}{l|lll|l|ll}
\hline
Dataset & Train & Valid & Test & \# of tokens & \# of rows & \# of columns \\
\hline
E2E & 42.1k & 4.7k & 4.7k & 24.90 & 4.58 & 2.00 \\
WikiTableText & 10.0k & 1.3k & 2.0k & 19.59 & 4.26 & 2.00 \\
WikiBio & 582.7k & 72.8k & 72.7k & 122.30 & 4.20 & 2.00 \\
\hline
\end{tabular}
\caption{Statistics of E2E, WikiTableText, and WikiBio datasets, including the number of instances in training, validation, and test sets, number of BPE tokens per instance, and number of rows per instance.}
\label{tab:dataset_statistics_2}
\end{table*}

\begin{table}[!htb]
\centering
\small
\begin{tabular}{ccc|c}
\hline
Train & Valid & Test & \# of tokens \\
\hline
3.4k & 727 & 728 & 351.05 \\
\hline
\end{tabular}
\begin{tabular}{c|ccc}
\hline
& \# of rows & \# of columns & \# of cells \\
\hline
Team & 2.71 & 4.84 & 6.56 (85.40\%) \\
Player & 7.26 & 8.75 & 22.63 (43.93\%) \\
\hline
\end{tabular}
\caption{Statistics of Rotowire dataset. The first table shows sizes of training, validation, and test sets, and the number of BPE tokens per instance. The second table shows the number of rows, the number of columns, and the number and ratio of non-empty cells.}
\label{tab:dataset_statistics_rotowire}
\end{table}

\section{Experiments}

\subsection{Datasets}

We make use of four existing datasets which are traditionally utilized for data-to-text: Rotowire \citep{wiseman2017challenges}, E2E \citep{novikova2017e2e}, WikiTableText \citep{bao2018table}, and WikiBio \citep{lebret2016neural}.
In each dataset, we filter out the content in the tables that does not appear in the texts. We plan to make the processed datasets publicly available for future research.
Table~\ref{tab:dataset_statistics_rotowire} gives the statistics of the Rotowire dataset and Table~\ref{tab:dataset_statistics_2} gives the statistics of the other three datasets.

\textbf{Rotowire} is from the sports domain. Each instance is composed of a text and two tables, where the text is a report of a basketball game and the two tables represent the scores of teams and players respectively (cf., Figure \ref{fig:rotowire_example}). Each table has column headers describing the types of scores, and row headers describing the names of teams or players.
The texts are long and may contain irrelevant information such as the performance of players in other games. Therefore, this is a challenging dataset.

\textbf{E2E} is from the restaurant domain. Each instance is a pair of short text and an automatically constructed table, where the text is a description of a restaurant, and the table has two columns with row headers summarizing the characteristics of the restaurant. %\red{The texts are short sentences and the tables lack diversity.} 
The tables are automatically constructed, where the texts in the tables are from a limited set and thus lack diversity.

\textbf{WikiTableText} is an open-domain dataset. Each instance includes a text and a table, where the text is a description and the table has a row and two columns with row headers collected from Wikipedia. 
The texts are short and contain information similar to that in the tables. 
%There exist multiple ``correct" tables given the same text.
%This makes the evaluation based on exact match slightly inaccurate.

\textbf{WikiBio} is extracted from the Wikipedia biography pages. Each instance consists of a text and a table, where the text is the introduction of Wikipedia page\footnote{The original dataset only uses the first sentence of the introduction. We use the entire introduction.} and the table is from the infobox of a Wikipedia page and has two columns with row headers. The input texts are usually long and contain more information than the tables. %, and thus it is necessary to summarize the key information into the tables. 
%Similar to WikiTableText, multiple ``correct'' tables exist in WikiBio.

\subsection{Procedure}

\textbf{Methods:} We conduct experiments with vanilla seq2seq and our method, as well as baselines. 

We know of no existing method that can be directly employed in text-to-table.
For each dataset, we first define the schemas based on the training data, then use an existing method of relation extraction (RE) or named entity extraction (NER) to extract information, and finally create tables based on the schemas and extracted information. We take it as the baseline for the dataset. No baseline can be applied to all four datasets. For RE, we use PURE, a state-of-the-art method~\citep{zhong-chen-2021-frustratingly}.
For NER, we use BERT~\citep{devlin2019bert}. %Table \ref{tab:baseline} summarizes the sub-tasks employed in the four datasets.

\begin{table*}[!htb]
\centering
\small
\begin{tabular}{l|l|ccc|ccc|ccc|c}
\hline
& \multirow{2}{*}{Model} & \multicolumn{3}{c|}{Row header F1} & \multicolumn{3}{c|}{Column header F1} & \multicolumn{3}{c|}{Non-header cell F1} & Err. \\
& & Exact & Chrf & BERT & Exact & Chrf & BERT & Exact & Chrf & BERT & rate \\
\hline \hline
\multirow{4}{*}{Team} & Sent-level RE & 85.28 & 87.12 & 93.65 & 85.54 & 87.99 & 87.53 & 77.17 & 79.10 & 87.48 & 0.00 \\
& Doc-level RE & 84.90 & 86.73 & 93.44 & 85.46 & 88.09 & 87.99 & 75.66 & 77.89 & 87.82 & 0.00 \\
& Vanilla seq2seq & 94.71 & 94.93 & 97.35
 & \textbf{86.07} & 89.18 & 88.90 & 82.97 & 84.43 & 90.62 & 0.49 \\
& Our method & \textbf{94.97} & \textbf{95.20} & \textbf{97.51} & 86.02 & \textbf{89.24} & \textbf{89.05} & \textbf{83.36} & \textbf{84.76} & \textbf{90.80} & 0.00 \\
\hline \hline
\multirow{4}{*}{Player} & Sent-level RE & 89.05 & 93.00 & 90.98 & 86.36 & 89.38 & 93.07 & 79.59 & 83.42 & 85.35 & 0.00 \\
& Doc-level RE & 89.26 & 93.28 & 91.19 & 87.35 & 90.22 & 97.30 & 80.76 & 84.64 & 86.50 & 0.00 \\
& Vanilla seq2seq & 92.16 & 93.89 & 93.60 & \textbf{87.82} & \textbf{91.28} & \textbf{94.44} & 81.96 & 84.19 & 88.66 & 7.40 \\
& Our method & \textbf{92.31} & \textbf{94.00} & \textbf{93.71} & 87.78 & 91.26 & 94.41 & \textbf{82.53} & \textbf{84.74} & \textbf{88.97} & 0.00 \\
\hline
\end{tabular}
\caption{Results of our method, vanilla seq2seq, and the baselines of doc-level RE and sent-level RE, on Rotowire. We show the F1 score based on exact match (Exact), chrf score (Chrf), and BERTScore (BERT) respectively.}
\label{tab:results_rotowire}
\end{table*}

\begin{table*}[!htb]
\centering
\begin{tabular}{lIlIcccIcccIc}
\hline
\multirow{2}{*}{Dataset} & \multirow{2}{*}{Model} & \multicolumn{3}{c|}{Row header F1} & \multicolumn{3}{c|}{Non-header cell F1} & Err. \\
& & Exact & Chrf & BERT & Exact & Chrf & BERT & rate \\
\hline \hline
\multirow{3}{*}{E2E} & NER & 91.23 & 92.40 & 95.34 & 90.80 & 90.97 & 92.20 & 0.00 \\
& Vanilla seq2seq & 99.62 & \textbf{99.69} & \textbf{99.88} & 97.87 & 97.99 & 98.56 & 0.00 \\
& Our method & \textbf{99.63} & \textbf{99.69} & \textbf{99.88} & \textbf{97.88} & \textbf{98.00} & \textbf{98.57} & 0.00 \\
\hline \hline
\multirow{3}{*}{WikiTableText} & NER & 59.72 & 70.98 & 94.36 & 52.23 & 59.62 & 73.40 & 0.00 \\
& Vanilla seq2seq & 78.15 & \textbf{84.00} & 95.60 & \textbf{59.26} & \textbf{69.12} & 80.69 & 0.41 \\
& Our method & \textbf{78.16} & 83.96 & \textbf{95.68} & 59.14 & 68.95 & \textbf{80.74} & 0.00 \\
\hline \hline
\multirow{3}{*}{WikiBio} & NER & 63.99 & 71.19 & 81.03 & 56.51 & 62.52 & 61.95 & 0.00 \\
& Vanilla seq2seq & \textbf{80.53} & \textbf{84.98} & \textbf{92.61} & 68.98 & \textbf{77.16} & 76.54 & 0.00 \\
& Our method & 80.52 & 84.96 & 92.60 & \textbf{69.02} & \textbf{77.16} & \textbf{76.56} & 0.00 \\
\hline
\end{tabular}
\caption{Results of our method, vanilla seq2seq, and the baseline of NER, on E2E, WikiTableText, and WikiBio. We show the F1 score based on exact match (Exact), chrf score (Chrf), and BERTScore (BERT) respectively.}
\label{tab:results_wiki}
\end{table*}

\textbf{Training:} 
For vanilla seq2seq and our method, we adopt Transformer~\citep{vaswani2017attention} as the model and fine-tune the models from BART-base. We also experiment with BART-large.
For RE and NER, we fine-tune the models from BERT-base-uncased.
All models are trained with Adam optimizer until convergence. %, and the hyper-parameters are tuned on the development sets. 
Hyper-parameters are shown in Appendix \ref{sec:appendix:hyperparameters}.
For the small datasets of Rotowire and WikiTableText, we run experiments five times with different random seeds and take the average of results to reduce variance. %\red{[check: not for baseline?]}

\textbf{Evaluation:} We evaluate the performance of a method based on (1) the number of correct headers and (2) the number of correct non-header cells. We adopt the F1 score as the evaluation measure. For each table, we compare the set of predicted results $\mathbf{y}$ against the set of ground-truth $\mathbf{y^*}$. Precision is defined as the percentage of the correctly predicted results among the predicted results, i.e., $P = \frac{1}{|\mathbf{y}|} \sum_{y\in\mathbf{y}} \max_{y^*\in\mathbf{y^*}} O(y, y^*)$. Recall is defined as the percentage of the correctly predicted results among the ground-truth, i.e., $R = \frac{1}{|\mathbf{y^*}|} \sum_{y^*\in\mathbf{y^*}} \max_{y\in\mathbf{y}} O(y, y^*)$. Finally, $F1=2/(1/P + 1/R)$. Here, $O(\cdot)$ denotes a way of similarity calculation. We consider three ways: exact match, chrf~\citep{DBLP:conf/wmt/Popovic15} and rescaled BERTScore~\citep{DBLP:conf/iclr/ZhangKWWA20}.
Exact match conducts an exact match between two texts.
Chrf calculates character-level n-gram similarity between two texts.
BERTScore calculates the similarity of BERT embeddings between two texts.
For non-header cells, we use not only the content but also the header(s) to ensure that the cell is on the right row (and column), and calculate the similarity $O(\cdot)$ as the product of header similarity and cell content similarity.\footnote{As shown in Figure \ref{fig:rotowire_example}, the tables in the dataset contain empty cells. The empty cells do not contain information. Therefore, we ignore the empty cells and only use the non-empty cells in the evaluation.}
We evaluate the measures of a generated table and then take the average on all tables.
This evaluation assumes that the order of rows and columns is not important.
We find that this assumption is applicable to the four datasets and many real-world scenarios.
% For example, in the table in Figure \ref{fig:rotowire_example}, permutation of the players or score types does not affect the contents of the table.}
We also evaluate the percentage of output sequences that cannot represent well-formulated tables, referred to as error rate.

\subsection{Results on Rotowire}

Table~\ref{tab:results_rotowire} shows the results on the Rotowire dataset. 
One can see that our method performs the best followed by vanilla seq2seq in terms of most of the measures, especially the F1 score on non-header cells. Both outperform the baselines of doc-level RE and sent-level RE. The RE baselines perform quite well, but they heavily rely on rules and cannot beat the seq2seq approach. Among them, the doc-level RE performs better than sent-level RE, because some information in Rotowire can only be extracted when the cross-sentence context is provided.
% 我觉得这个更多是这个数据集本身的原因，而不是in general的IE就需要在document-level执行。information extraction needs to be conducted in broader contexts.

We implement two baselines of RE, namely doc-level RE and sent-level RE.  We take team names, player names, and numbers of scores as entities and take types of scores as relations. % \red{An example relation tuple is (Al Horford, Assists, 5), where ``Al Horford'' is the subject entity, ``5'' is the object entity, and Assists is one of the pre-defined relation types.}
% There are $38$ relations in total. 
Sent-level RE predicts the relations between entities within each sentence.
Doc-level RE predicts the relations between entities within a window (the window size is $12$ entities) and uses the approximation model proposed by \citet{zhong-chen-2021-frustratingly} to speed up inference.

\begin{table*}[!htb]
\centering
\begin{tabular}{ccc|ccccc}
\hline
Pre & TC & TRE & Rotowire/Team & Rotowire/Player & E2E & WikiTableText & WikiBio \\
\hline
\xmark & \xmark & \xmark & 28.05 & 7.75 & 94.45 & 46.37 & 67.51 \\
\xmark & \cmark & \cmark & 30.61 & 10.67 & 95.53 & 47.13 & 67.43 \\
\hline
\cmark & \xmark & \xmark & 82.97 & 81.96 & 97.87 & 59.26 & 68.98 \\
\cmark & \cmark & \xmark & 83.09$^\ddagger$ & 82.24$^\ddagger$ & \textbf{97.88} & \textbf{59.29}$^\dagger$ & 68.98 \\
\cmark & \xmark & \cmark & 83.30$^\dagger$ & 82.50$^\ddagger$ & 97.87 & 59.12 & \textbf{69.02} \\
\cmark & \cmark & \cmark & \textbf{83.36}$^\dagger$ & \textbf{82.53}$^\ddagger$ & \textbf{97.88} & 59.14  & \textbf{69.02} \\
\hline
\end{tabular}
\caption{Results of ablation study on our method by excluding pre-trained language model (Pre), table constraint (TC) and table relation embeddings (TRE). We report F1 for non-header cells based on exact match. We conduct a significance test to check whether the performance is significantly better than vanilla seq2seq with pre-trained language models (i.e., with Pre but without TC or TRE). $^\dagger$ and $^\ddagger$ represent $p<0.05$ and $p<0.01$ respectively.}
\label{tab:results_ablation}
\end{table*}
\begin{table*}[!htb]
\centering
\begin{tabular}{l|c@{\hspace{\tabcolsep}}c@{\hspace{\tabcolsep}}c@{\hspace{\tabcolsep}}c@{\hspace{\tabcolsep}}c}
\hline
Method & Rotowire/Team & Rotowire/Player & E2E & WikiTableText & WikiBio \\
\hline
Vanilla seq2seq (BART base) & 82.97 & 81.96 & 97.87 & 59.26 & 68.98 \\
Our method (BART base) & 83.36 & 82.53 & 97.88 & 59.14  & 69.02 \\
Vanilla seq2seq (BART large) & \textbf{86.31} & 86.59 & \textbf{97.94} & \textbf{62.71} & 69.66 \\
Our method (BART large) & \textbf{86.31} & \textbf{86.83} & 97.90 & 62.41 & \textbf{69.71} \\
\hline
\end{tabular}
\caption{Results of our method and vanilla seq2seq with base and large BART models on all four datasets.}
\label{tab:results_BART}
\end{table*}
\begin{figure*}[!htb]
\centering
\includegraphics[width=0.8\linewidth]{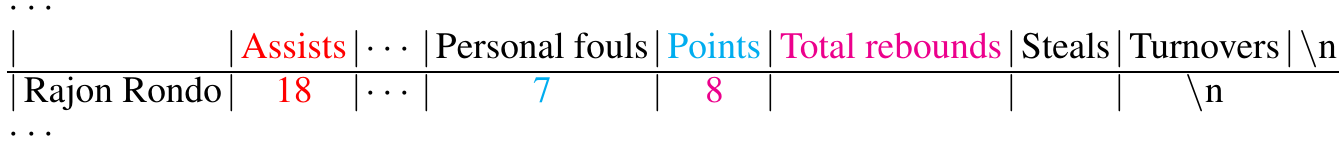}
\caption{A bad case generated by vanilla seq2seq. The assists, points and total rebounds of Rajon Rondo should be $18$, $7$ and $8$ respectively. The model generates one less column between ``Assists'' and ``Personal fouls''.}
\label{fig:rotowire_bad_case}
\end{figure*}

\subsection{Results on E2E, WikiTableText and WikiBio}

Table~\ref{tab:results_wiki} shows the results of our method, vanilla seq2seq, and the baseline of NER on E2E, WikiTableText, and WikiBio. Again, the seq2seq approach outperforms the baseline. Our method and vanilla seq2seq are comparable, because the table structures in the three datasets are very simple (there are only two columns in the tables), and the use of the two techniques does not further improve the performances. The NER baseline has high precision but low recall, mainly because NER can only make the right decision when it is clear.

We implement the baseline of NER in the following way. We view the non-head cells in the tables as entities and their row headers as entity types. % \red{For example, suppose a row in a table has a header ``Name'' and a non-header cell ``The Vault''. We take ``The Vault'' as an entity with type ``Name'', where ``Name'' is one of the pre-defined entity types.} \red{[del]}
In training, we match the non-head cells into the texts and take them as ``entities'' in the texts. Only a proportion of the non-header cells can be matched into the texts ($85\%$ for E2E, $74\%$ for WikiTableText, and $69\%$ for WikiBio).

\subsection{Additional Study}

We carry out an ablation study on our method. Specifically, we exclude pre-trained language model, table constraint (TC), and table relation embeddings (TRE) from our method.
Note that our method without TC and TRE is equivalent to vanilla seq2seq. Table~\ref{tab:results_ablation} gives the results on the four datasets.

It can be seen that the use of both TC and TRE can significantly improve the performance on Rotowire, which indicates that our method is particularly effective when the tables are large with many rows and columns. There are no significant improvements on E2E, WikiTableText, and  WikiTableText, apparently because the formulation of tables is easy for the three datasets. Therefore, we conclude that the two techniques of TC and TRE are helpful when the task is difficult.

The use of pre-trained language model can boost the performance on all datasets, especially on Rotowire and WikiTableText. This indicates that pre-trained language model is particularly helpful when the task is difficult and the size of training data is small. %Further note that our method without a pre-trained language model can work even better than vanilla seq2seq on Rotowire, E2E, and WikiTableText.

We observe that vanilla seq2seq makes more formatting errors than our method, especially on player tables in Rotowire that have a large number of columns. It indicates that for vanilla seq2seq, it is difficult to keep track of the columns in each row and make alignments with the column headers. In contrast, the two techniques of our method can help effectively cope with the problem. Figure \ref{fig:rotowire_bad_case} shows a bad case of vanilla seq2seq, where the model correctly infers the column of ``assists'' but fails to infer the columns of ``personal fouls'', ``points'', and ``total rebounds'' for the row of ``Rajon Rondo''. In contrast, our method can successfully handle the case, because TC can eliminate the incorrectly formatted output, and TRE can make correct alignments with the column headers.

We also investigate the effect of the scale of pre-trained language model BART. We use both BART-base and BART-large and conduct fine-tuning on top of them for vanilla seq2seq and our method. Table~\ref{tab:results_BART} gives the results on the four datasets. The results show that the use of BART-large can further boost the performances on all four datasets, indicating that it is better to use larger pre-trained models when computation cost is not an issue.

\subsection{Discussions}
\label{sec:discussions}

We analyze the experimental results on the four datasets and identify five challenging issues.

(1) Text Diversity:
Extraction of the same content from different expressions is one challenge. For example, the use of synonyms is very common in Rotowire. The team of ``Knicks'' is often referred to as ``New York'', its home city. Identification of the same entities from different expressions is needed in the task.

(2) Text Redundancy: 
There are cases such as those in WikiBio, in which the texts contain much redundant information. This poses a challenge to the text-to-table model to have a strong ability in summarization. It seems that the seq2seq approach works well to some extent but further improvement is undoubtedly necessary.

(3) Large Table: 
The tables in Rotowire have large numbers of columns, and the extraction from them is challenging even for our method of using TC and TRE. 

(4) Background Knowledge: 
WikiTableText and WikiBio are from open domain. 
Thus, performing text-to-table on such kind of datasets require the use of much background knowledge. A possible way to address this challenge is to use more powerful pre-trained language models or external knowledge bases. % The use of more powerful pre-trained language models can be further explored in the future.

(5) Reasoning: 
Sometimes the information is not explicitly presented in the text, and reasoning is required to conduct correct extraction. 
For example, an article in Rotowire reports a game between the two teams ``Nets'' and ``Wizards''.
From the sentence: ``The Nets seized control of this game from the very start, opening up a 31 - 14 lead after the first quarter'', humans can infer that the point of ``Wizards'' is $14$, which is still difficult for machines.

\section{Conclusion}

We propose employing text-to-table as a new way of information extraction (IE), which extracts information of interest from the input text and summarizes the extracted information in tables.  The advantage of the approach is that one can easily conduct information extraction from either short texts or long texts to create simple tables or complex tables without explicitly defining the schemas. Text-to-table can be viewed as an inverse problem of table-to-text. We formalize text-to-table as a sequence-to-sequence problem on top of a pre-trained model. We further propose an improved method using a seq2seq model and table constraint and table relation embeddings techniques.  We conduct experiments on four datasets derived from existing table-to-text datasets. The results demonstrate that our proposed approach outperforms existing methods using conventional IE techniques. We further analyze the challenges of text-to-table for future study. The issues include diversity of text, redundancy of text, large table, background knowledge, and reasoning.

\bibliography{anthology,custom}
\bibliographystyle{acl_natbib}

\appendix

\section{Hyper-parameters}
\label{sec:appendix:hyperparameters}

We list the hyper-parameters of the pre-trained models in Table~\ref{tab:hyperpara_pretrained}. The training hyper-parameters for BART-base model in vanilla seq2seq and our method are listed in Table~\ref{tab:hyperpara_training}.

\begin{table*}[!htbp]
    \centering
    \begin{tabular}{l|l|cccc}
    \hline
    Pre-trained model & Methods & layers & hidden dim. & heads & parameters \\
    \hline
    BART-base & Vanilla seq2seq and ours & 12 & 768 & 16 & 139M \\
    BART-large & Vanilla seq2seq and ours & 24 & 1024 & 16 & 406M \\
    BERT-base-uncased & NER and RE & 12 & 768 & 12 & 110M \\
    \hline
    \end{tabular}
    \caption{The hyper-parameters of the pre-trained models in our experiments. We list the number of layers, hidden dimensions (hidden dim.), heads, and parameters. BART-base and BART-large are used for vanilla seq2seq and our method, while BERT-base-uncased is used for the baselines of RE and NER.}
    \label{tab:hyperpara_pretrained}
\end{table*}

\begin{table}[!htbp]
    \centering
    \small
    \begin{tabular}{l|cccc}
    \hline
    & warmup & total upd. & lr & bsz \\
    \hline
    Rotowire & 400 & 8000 & 3e-05 & 4096 \\
    E2E & 400 & 8000 & 1e-05 & 4096 \\
    WikiTableText & 2000 & 8000 & 1e-04 & 4096 \\
    WikiBio & 4000 & 40000 & 1e-04 & 4096 \\
    \hline
    \end{tabular}
    \caption{The training hyper-parameters for BART-base model on all four datasets. We list the warmup updates (warmup), total updates (total upd.), learning rate (lr), and batch size (bsz, in terms of how many tokens per batch).}
    \label{tab:hyperpara_training}
\end{table}

\section{Table Constraint Algorithm}
\label{sec:appendix:constrained_decoding}

The pseudo-codes for table constraint are in Algorithm \ref{alg:constrained_decoding}.

\begin{algorithm2e}[!htb]
\DontPrintSemicolon
  
  \SetKwBlock{RepeatForever}{repeat}{}
  \SetKw{Break}{break}
  
  \KwInput{$\mathbf{x}=[x_1,x_2,\cdots,x_{|\mathbf{x}|}]$}
%   \tcc{Input sequence $\mathbf{x}=[x_1,x_2,\cdots,x_{|\mathbf{x}|}]$}
  \KwOutput{$\mathbf{y}=[y_1,y_2,\cdots,y_{|\mathbf{y}|}]$}
%   \tcc{Output sequence $\mathbf{y}=[y_1,y_2,\cdots,y_{|\mathbf{y}|}]$}
%   \KwData{Testing set $x$}

    $\mathbf{y} \leftarrow []$\;
    \;
            
        \Repeat {\upshape $\mathbf{y}_{|\mathbf{y}|} = \nline{}$ or $\mathbf{y}_{|\mathbf{y}|} = \eos{}$ \label{alg:constrained_decoding:repeat_first_row_end}} { \label{alg:constrained_decoding:repeat_first_row_start}
            \tcc{generates the first row: only allows generation of \nline{} or \eos{} after \sep{} }
            $p(\cdot) \leftarrow $ \FuncSty{seq2seq}($\mathbf{x}$, $\mathbf{y}$) \;
            % \tcc{Use the seq2seq model to predict the distribution of the next output token}
            \uIf{\upshape $\mathbf{y}_{|\mathbf{y}|} \neq \sep{}$ }{
            $p\left( \nline{} \right) \leftarrow 0,\ p\left( \eos{} \right) \leftarrow 0$ \;
            }
            
            $\mathbf{y}$.append$\left(\FuncSty{decode}\left(p\right)\right)$\;
        }
            \uIf{\upshape $\mathbf{y}_{|\mathbf{y}|} = \eos{}$ }{
            % \uIf{ $\mathbf{y}_{|\mathbf{y}|} = \eos{}$ }{
                \Return $\mathbf{y}$ \;
            }
            $n_c \leftarrow$ number of cells of the first row\; \label{alg:constrained_decoding:calc_nc}
            \;
            
            \RepeatForever{
                \tcc{generates the next rows: each row contains exactly $n_c$ cells }
                
        \Repeat {\upshape $\mathbf{y}_{|\mathbf{y}|} = \nline{}$ or $\mathbf{y}_{|\mathbf{y}|} = \eos{}$ \label{alg:constrained_decoding:repeat_next_row_end}}{ \label{alg:constrained_decoding:repeat_next_row_start}
            \tcc{generates a row}
            $p(\cdot) \leftarrow $ \FuncSty{seq2seq}($\mathbf{x}$, $\mathbf{y}$) \;
            \uIf{\upshape current row has $n_c$ columns }{
                $p(t) \leftarrow 0,\ \forall t \neq \eos{} \textmd{ and } t \neq \nline{}$\;
            } \uElse{
                $p\left( \nline{} \right) \leftarrow 0,\ p\left( \eos{} \right) \leftarrow 0$ \;
            }
            $\mathbf{y}$.append$\left(\FuncSty{decode}\left(p\right)\right)$\;
                }
                \uIf{\upshape $\mathbf{y}_{|\mathbf{y}|} = \eos{}$ }{
                    \Return $\mathbf{y}$ \;
                }
        }
\caption{Decoding using table constraint. \eos{}, \sep{}, and \nline{} denote the end of sentence, separation token, and new-line token respectively. Seq2seq denotes the seq2seq model. Decode denotes the decoding algorithm such as beam search and greedy search.}
\label{alg:constrained_decoding}
\end{algorithm2e}

\section{Our Method with Multiple Tables}
\label{sec:appendix:multitable}

Our method is able to generate the output containing multiple tables. For example, in Rotowire dataset, the output data contains two tables representing the scores of teams and players respectively. In this section, we illustrate how our method works for Rotowire dataset as a special case.

To represent the tables with a sequence, we use captions as delimiters. For Rotowire, as shown in Figure \ref{fig:rotowire_example}, the first table is the team table, and its caption is ``Team:''. The second table is the player table, and its caption is ``Player:''. Let $\mathbf{t}^\text{team}$ and $\mathbf{t}^\text{player}$  denote the table and player tables respectively. Therefore, the sequence representation is ``Team: \nline{} $\mathbf{t}^\text{team}$ \nline{} Player: \nline{} $\mathbf{t}^\text{player}$''. Although this example contains only two tables, the seq2seq model can generate any number of tables during the generation process until it reaches \eos{} and stops decoding. Therefore, there is no need to pre-determine the number of output tables.

For table constraint (TC), we only use TC when the seq2seq model is generating a table. When generating a caption, we do not pose any constraints to the decoding process.
Since the captions do not start with the separation token \sep{}, if the current line starts with the separation token \sep{}, then the model is generating a table. Otherwise, it is generating a caption.

For table relation embeddings (TRE), we calculate the relation vectors separately for each table. However, the parameters including the row relation embeddings (i.e., $\tau_r^K$ and $\tau_r^V$) and the column relation embeddings (i.e., $\tau_c^K$ and $\tau_c^V$) are shared among the tables.

\section{Information Extraction Baselines}

\subsection{Relation Extraction}
\label{sec:appendix:baseline:re}

\begin{figure*}[htb]
\centering
\includegraphics[width=\linewidth]{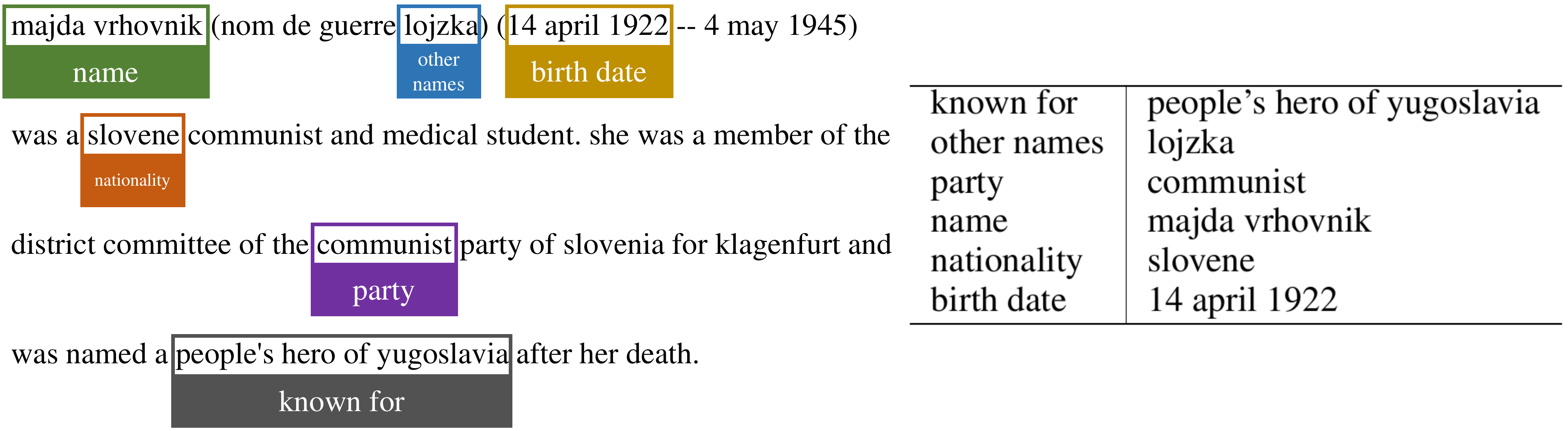}
\caption{An example of NER data on WikiBio dataset. Each row header is an entity type, and each non-header cell is an entity.}
\label{fig:wiki_ner_example}
\end{figure*}
\begin{figure}[htb]
\centering
\includegraphics[width=0.8\linewidth]{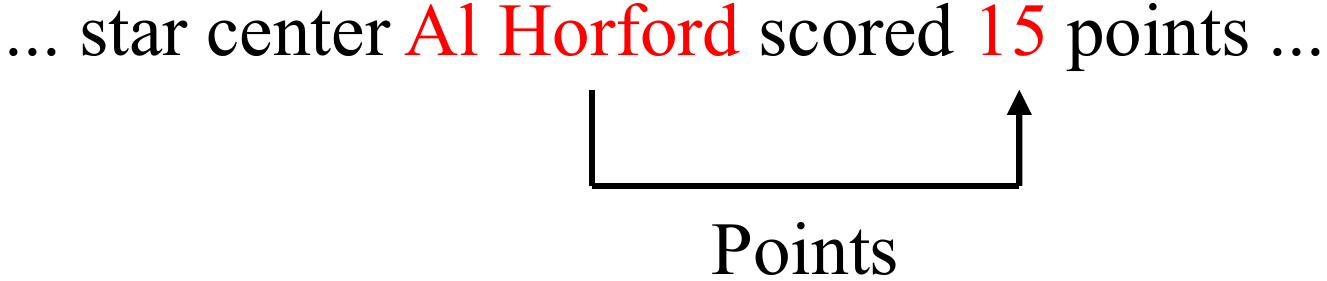}
\caption{An example of RE data from Rotowire dataset. ``Al Horford'' and ``15'' are entities, and ``Points'' is the relation type.}
\label{fig:rotowire_re_example}
\end{figure}
\begin{figure}[!tbh]
\centering
\includegraphics[width=0.95\linewidth]{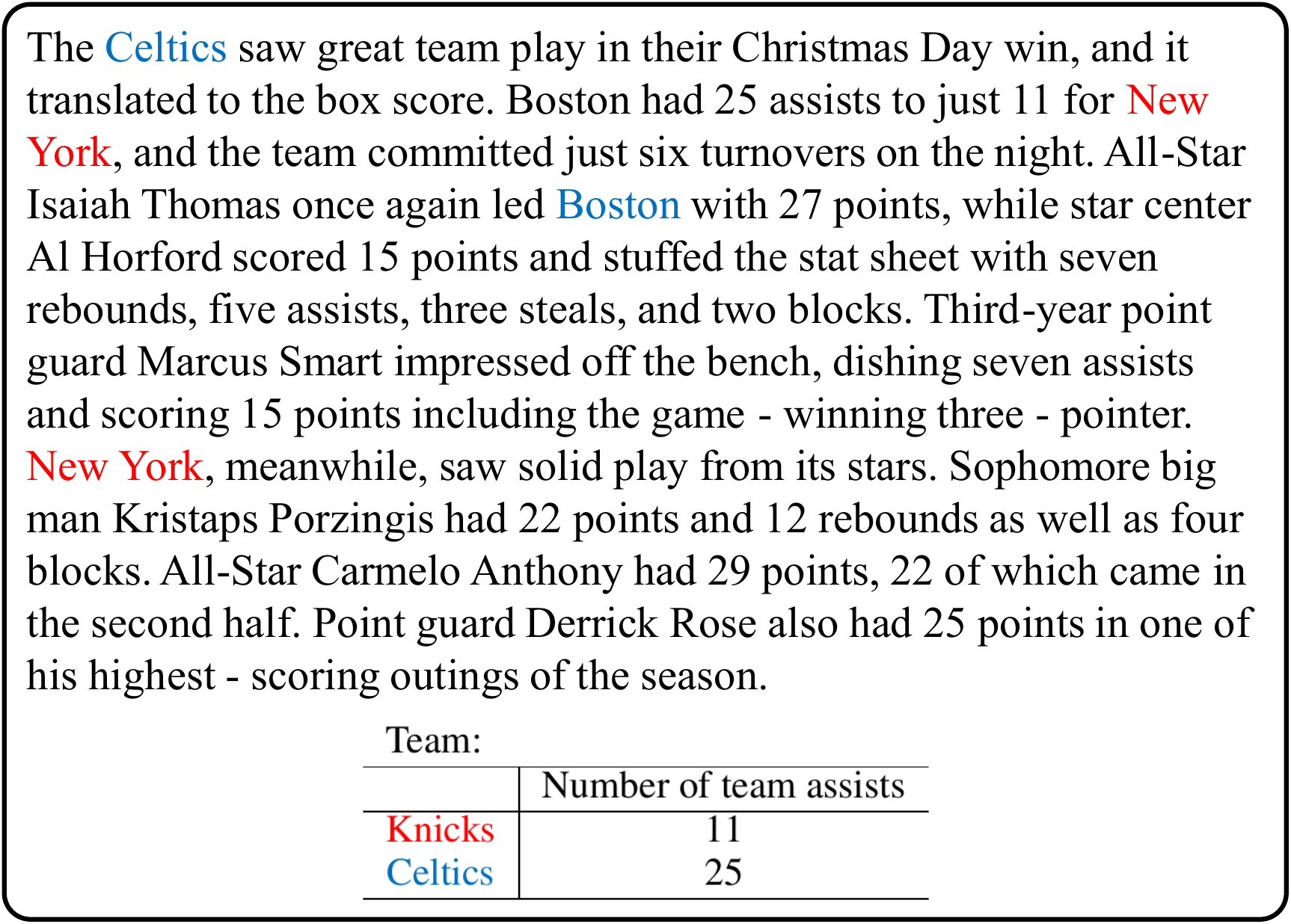}
\caption{Illustration of the use of synonyms for the example in Figure \ref{fig:rotowire_example}. The red color denotes the team of ``Knicks'', which is often referred to as ``New York'', its home city. The blue color denotes the team of ``Celtics'', which is often referred to as ``Boston'', its home city.}
\label{fig:rotowire_synonym}
\end{figure}

To use relation extraction (RE) as our baseline for Rotowire dataset, we take team names, player names, and numbers of scores as entities and take types of scores as relations. An example relation is shown in Figure \ref{fig:rotowire_re_example}, which can be represented as a relation tuple (Al Horford, Points, 15). ``Al Horford'' is the subject entity, ``15'' is the object entity, and ``Points'' is one of the pre-defined relation types.
There are $38$ relations in total.

To create synthetic training data, we match the player names, team names and score numbres to the texts. We adapt the rules provided by \citet{wiseman2017challenges} which is able to conduct fuzzy match.

\subsection{Named Entity Recognition}
\label{sec:appendix:baseline:ner}

\begin{figure*}[tbh]
\centering
\includegraphics[width=\linewidth]{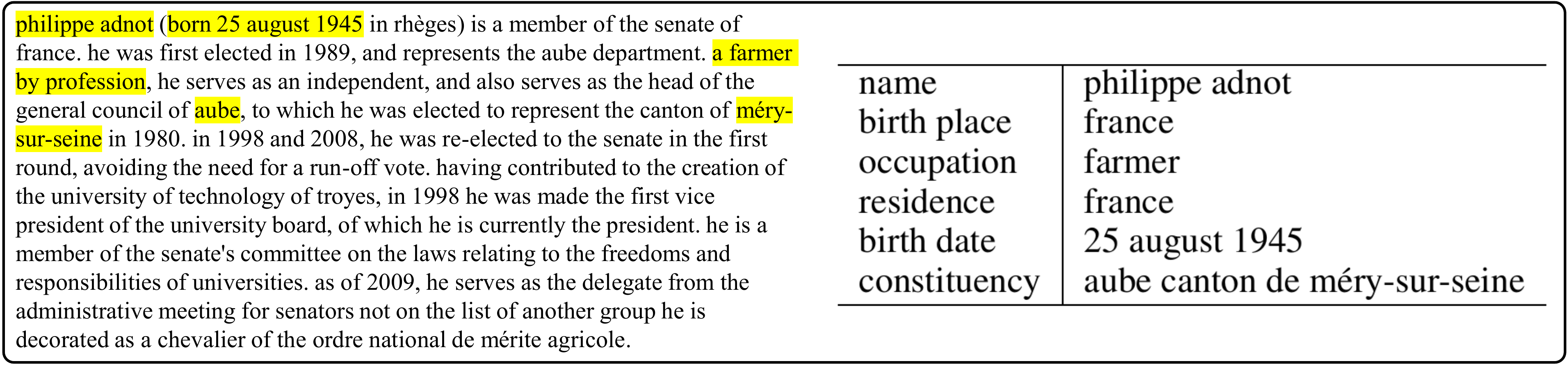}
\caption{An example from WikiBio dataset to illustrate the challenges of text redundancy and background knowledge. Only the highlighted information is captured in the output table. Other information such as the experience of Philippe Adnot is redundant. Moreover, the system should have background knowledge about the French political system to extract information about the constituency of Philippe Adnot.}
\label{fig:wikibio_redundancy}
\end{figure*}

We use named entity recognition (NER) as our baseline for E2E, WikiTableText, and WikiBio datasets.
Specifically, since each table is a two-column table with a header column, we consider the row header as entity type and the non-header cells as entity mentions. An example is shown in Figure \ref{fig:wiki_ner_example}.
For the row with a header ``name'' and a non-header cell ``majda vrhnovnik'', we take ``majda vrhnovnik'' as an entity with the type ``name''. Here, ``name'' is one of the pre-defined entity types.
We collect all headers in the training set to collect the entity types.
We have $7$ entity types for E2E, $2262$ entity types for WikiTableText, and $2272$ entity types for WikiBio.

To create synthetic training data, we match the contents of non-header cells to the texts.
However, the data is usually paraphrased or even abstracted from the text, so not all non-header cells can be matched to the text.
We match $85\%$ non-header cells for E2E, $74\%$ for WikiTable, and $69\%$ for WikiBio.

\section{Detailed Cases for Challenges}
\label{sec:appendix:discussions}

In this section, we provide cases for the challenges discussed in Section \ref{sec:discussions}.

\begin{figure*}[!tbh]
\centering
\includegraphics[width=\linewidth]{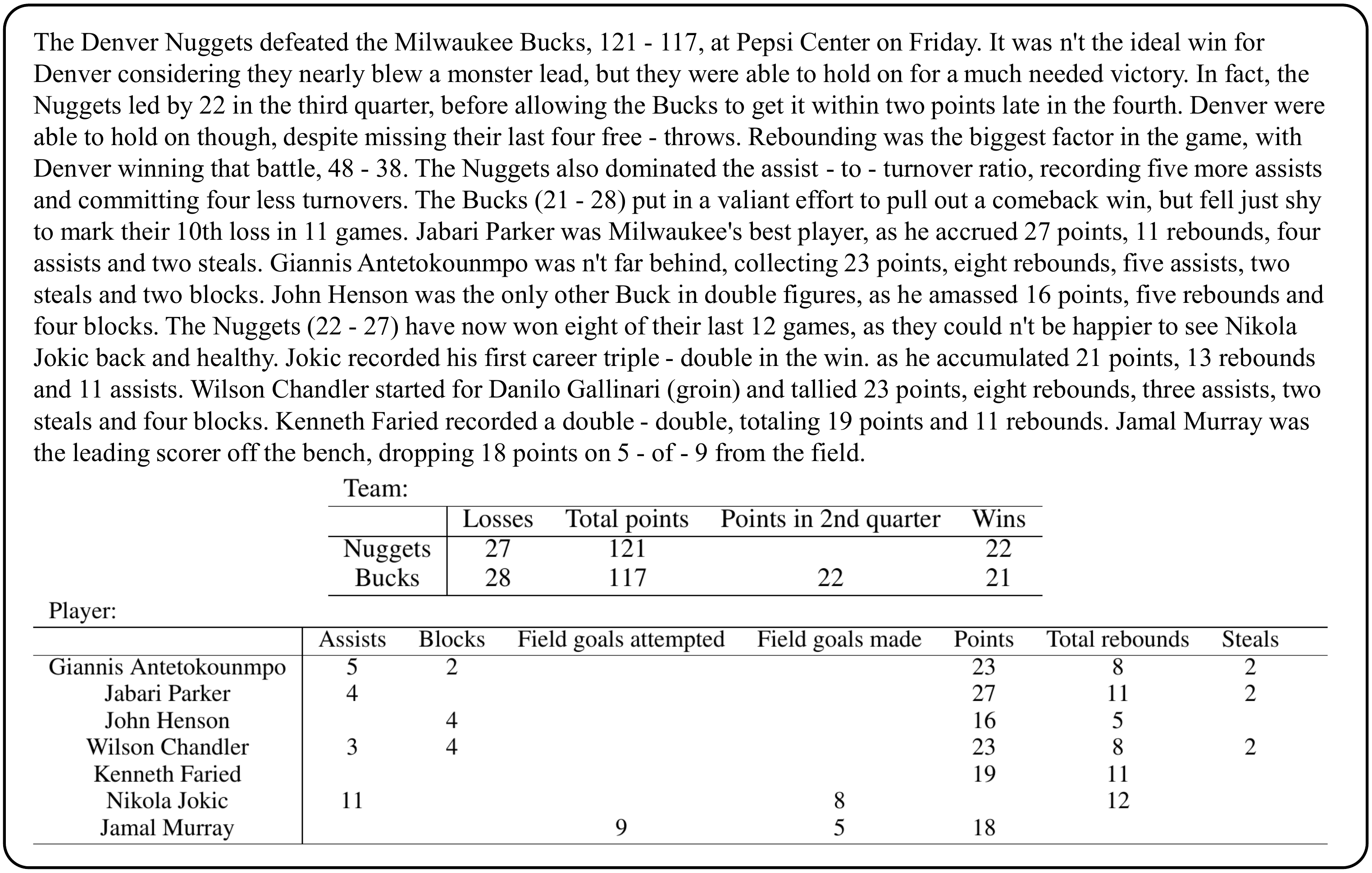}
\caption{The team table has 3 rows and 4 columns, and the player table has 8 rows and 8 columns.}
\label{fig:rotowire_big_table}
\end{figure*}

(1) Text Diversity:
Extraction of the same content from different expressions is one challenge.
For example, the use of synonyms is very common in Rotowire. Figure \ref{fig:rotowire_synonym} illustrates the use of synonyms for the example in Figure \ref{fig:rotowire_example}. The team of ``Knicks'' is often referred to as ``New York'', its home city. Similarly, ``Celtics'' is often referred to as ``Boston'', its home city. 
Identification of the same entities from different expressions is needed in the task.

\begin{figure}[htb]
\centering
\includegraphics[width=\linewidth]{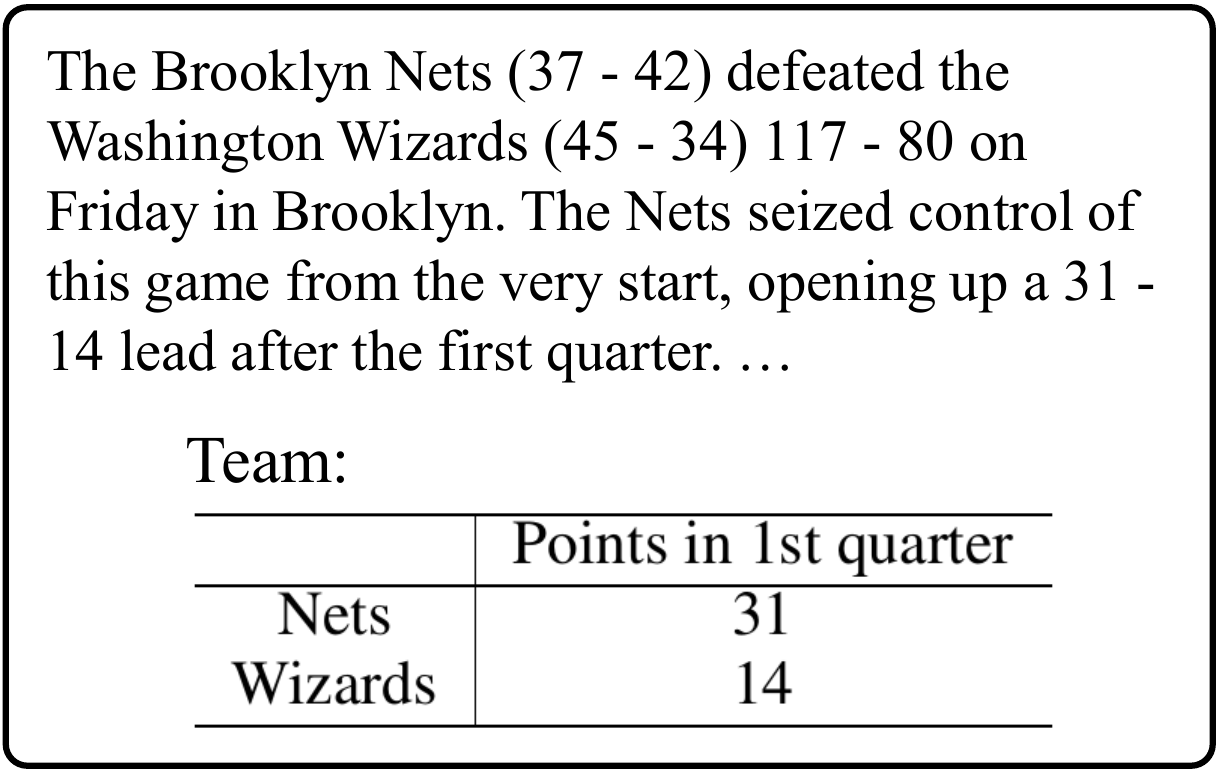}
\caption{An example from Rotowire which requires reasoning to perform information extraction. The article in Rotowire reports a game between the two teams ``Nets'' and ``Wizards''.
From the sentence: ``The Nets seized control of this game from the very start, opening up a 31 - 14 lead after the first quarter'', humans can infer that the point of ``Wizards'' is $14$, which is still difficult for machines.}
\label{fig:rotowire_reasoning}
\end{figure}

(2) Text Redundancy: 
There are cases such as those in WikiBio, in which the texts contain much redundant information. An example is shown in Figure \ref{fig:wikibio_redundancy}, where only the highlighted information is captured in the output table. Other information such as the experience of Philippe Adnot is redundant. This poses a challenge to the text-to-table model to have a strong ability in summarization. It seems that the seq2seq approach works well to some extent but further improvement is undoubtedly necessary.

(3) Large Table: 
The tables in Rotowire have large numbers of columns, so extraction from them is challenging even for our method of using TC and TRE. As presented in Table \ref{tab:dataset_statistics_rotowire}, team tables have 2.71 rows and 4.84 columns on average, and player tables have 7.26 rows and 8.75 columns on average. An example is shown in Figure \ref{fig:rotowire_big_table}, where the team table has 3 rows and 4 columns, and the player table has 8 rows and 8 columns. %\red{[check: even bigger?]}

(4) Background Knowledge: 
WikiTableText and WikiBio are from open domain. 
Thus, performing text-to-table on such kind of datasets require the use of much background knowledge.
Also in Figure \ref{fig:wikibio_redundancy}, the extraction system should have background knowledge about the French political system in order to extract information about the constituency of Philippe Adnot. A possible way to address this challenge is to use more powerful pre-trained language models or external knowledge bases.

(5) Reasoning: 
Sometimes the information is not explicitly presented in the text, and reasoning is required to conduct correct extraction. 
For example, as shown in Figure \ref{fig:rotowire_reasoning}, an article in Rotowire reports a game between the two teams ``Nets'' and ``Wizards''.
From the sentence: ``The Nets seized control of this game from the very start, opening up a 31 - 14 lead after the first quarter'', humans can infer that the point of ``Wizards'' is $14$, which is still difficult for machines.

\end{document}